\documentclass[letterpaper]{article} 
\usepackage[preprint]{aaai2027}  
\usepackage[hyphens]{url}  
\usepackage{graphicx} 
\usepackage{natbib}  
\usepackage{caption} 
\usepackage{algorithm}
\usepackage{algorithmic}

\usepackage{newfloat}
\usepackage{listings}
\DeclareCaptionStyle{ruled}{labelfont=normalfont,labelsep=colon,strut=off} 
\floatstyle{ruled}
\newfloat{listing}{tb}{lst}{}
\floatname{listing}{Listing}

\usepackage{booktabs}

\usepackage{bbding}
\usepackage{makecell}

\title{DashArena: Benchmarking LLMs on Interactive Analytic Dashboard Generation}

\author {
    Xiaotong Wang\textsuperscript{\rm 1},
    Dazhen Deng\textsuperscript{\rm 2}\corresponding
}
\affiliations {
    \textsuperscript{\rm 1}State Key Lab of CAD\&CG, Zhejiang University, China\\
    \textsuperscript{\rm 2}School of Software Technology, Zhejiang University, China\\
    xtwangcs@zju.edu.cn, dengdazhen@zju.edu.cn
}

\begin{document}

\maketitle

\begin{abstract}
Analytic dashboards combine coordinated views and interactions for data exploration and decision-making. Recent models can generate them from data and natural-language goals, but evaluating their usefulness remains difficult. Dashboard generation is open-ended, and neither static appearance nor successful execution alone captures analytical support and interaction quality.
We introduce DashArena, to our knowledge the first benchmark for open-ended, task-grounded generation of interactive analytic dashboards. Its key innovation is to require each system to generate both a dashboard and a replayable interaction trajectory. A browser executor replays the trajectory and turns the system's intended analytical workflow into reproducible visual and execution evidence. A VLM judge compares candidates using this evidence, and Bradley--Terry aggregation produces the leaderboard. We further distill the judge into the open-weight DashJudge-8B.
Human evaluations show that DashJudge-8B effectively reproduces human judgments and ablations show that interaction evidence improves judge agreement. Experiments with frontier models reveal persistent rendering, analytical, and interaction failures. Together, these results show that realistic dashboard generation remains challenging and that interaction-aware evaluation captures failures missed by static or execution-only checks.
\end{abstract}


\section{Introduction}

Analytic dashboards are interactive interfaces that summarize data while supporting exploratory analysis via coordinated multiple views~\cite{bach2022dashboard, sarikaya2018we, few2006information}.
Authoring them requires decisions about data transformation, visual encoding, multi-view organization, and interaction design~\cite{setlur2023heuristics, srinivasan2024dashboard}.
Recent LLMs can generate executable applications and analyze tabular data, making dashboard generation increasingly feasible~\cite{zhou2024webarena, hu2024infiagent}.

Evaluating this capability is difficult because dashboard generation is open-ended.
Multiple layouts, chart selections, and interaction designs may validly support the same analytical goal, so matching a single human reference or a fixed component checklist can penalize reasonable alternatives.
Static screenshots reveal visual quality but not whether interactions work.
Conversely, programmatic execution checks can detect broken actions but cannot determine whether a functional dashboard answers the task well.
With the rise of GUI agents~\cite{nguyen2025gui}, some benchmarks have proposed using them to explore web applications and evaluate their functionality~\cite{tran2026vibe}.
However, the evaluation result would depend on the agent's ability to discover an unfamiliar interface, and a missed feature would be indistinguishable from an absent feature.
Such exploration is also expensive and difficult to reproduce.
A useful benchmark must therefore preserve design freedom, obtain reproducible evidence about how each artifact is intended to be used, and leave semantic quality assessment to a separate evaluator.

We present \textbf{DashArena}, a benchmark for open-ended analytic dashboard generation in a controlled single-file web setting.
We use high-quality human-authored dashboards as task seeds and ask a VLM to observe their pages, data schema, and interaction features extracted from code, then express the underlying user context, analytical goals, and desired interactions as a natural-language task.
The task describes the intended analysis instead of requiring reproduction of the original dashboard.
For each task, a candidate LLM first generates a web dashboard and then, with access to its own implementation, produces a structured interaction trajectory describing a meaningful analytical walkthrough.
Following the replayable-demonstration principle of GameCraft-Bench~\cite{luo2026gamecraft}, we treat this trajectory as executable documentation supplied with the generated artifact.
It provides a concise demonstration of the analysis paths it claims to support.
Our browser executor independently replays every declared action and records whether it succeeds and which charts, controls, and text components change.
A VLM judge can therefore evaluate the dashboard's quality based on the task, screenshots, and execution evidence without needing to explore the interface itself.

\begin{figure}[t]
    \centering
    \includegraphics[width=\columnwidth]{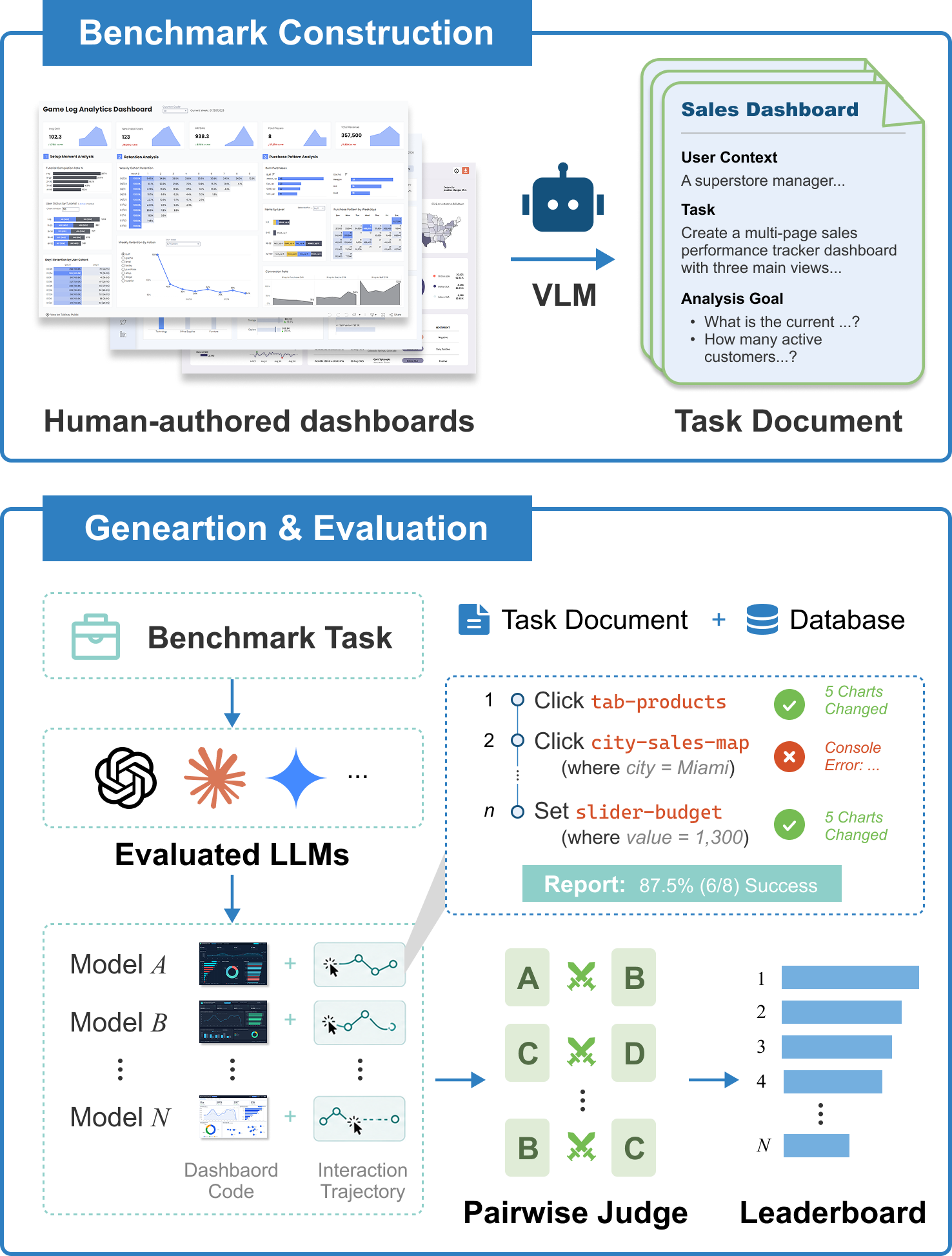}
    \caption{DashArena overview. Human-authored dashboards provide realistic task seeds. Each evaluated model generates a dashboard and a trajectory; browser execution turns that trajectory into deterministic evidence. The VLM judge compares anonymous candidates, and task-clustered Bradley--Terry aggregation produces the final leaderboard.}
    \label{fig:pipeline}
\end{figure}

DashArena evaluates candidates pairwise.
A VLM judge compares two anonymous dashboards and their trajectories, using screenshots and execution reports to predict their relative quality.
We first instantiate this protocol with Claude Opus 4.6 as a proprietary teacher and collect its preferences over task-stratified candidate pairs.
We then fine-tune DashJudge-8B, a VLM judge based on Qwen3-VL-8B-Instruct, using 255 independent teacher-labeled pairs with a data augmentation strategy that swaps candidate order and reverses the label.
It enables reproducible, high-throughput evaluation on local hardware.
We ran the evaluations and aggregated comparisons with a Bradley--Terry model~\cite{bradley1952rank} to produce a final leaderboard.

Table~\ref{tab:benchmark-comparison} positions DashArena relative to existing visualization benchmarks.
nvBench and nvBench 2.0 evaluate on natural language to visualization (NL2VIS) tasks, with the latter explicitly modeling controlled ambiguity~\cite{luo2021nvbench,luo2025nvbench}.
VisEval additionally evaluates for chart readability through a VLM judge~\cite{chen2024viseval}.
All three benchmarks focus on single-chart generation.
VisJudge-Bench evaluates the quality of static visualization images including dashboards, but does not consider interactivity and task completion~\cite{xie2026visjudge}.
To our knowledge, DashArena is the first benchmark for open-ended, task-grounded generation of interactive analytic dashboards. It is also the first dashboard-generation benchmark to use generator-authored, replayable interaction trajectories as evidence for runtime and interaction evaluation.

\begin{table*}[t]
\centering
\small
\setlength{\tabcolsep}{4pt}
\begin{tabular}{@{}ccccccc@{}}
\toprule
Benchmark & Primary artifact & \makecell[c]{Task-grounded\\evaluation} & \makecell[c]{Open-ended or\\ambiguous request} & \makecell[c]{Multi-view\\dashboard} & \makecell[c]{Visual quality\\evaluation} & \makecell[c]{Executed\\interactions} \\
\midrule
nvBench~\cite{luo2021nvbench} & Code-only & \Checkmark & \XSolidBrush & \XSolidBrush & \XSolidBrush & \XSolidBrush \\
nvBench 2.0~\cite{luo2025nvbench} & Code-only & \Checkmark & Controlled & \XSolidBrush & \XSolidBrush & \XSolidBrush \\
VisEval~\cite{chen2024viseval} & Code + image & \Checkmark & \XSolidBrush & \XSolidBrush & \Checkmark & \XSolidBrush \\
VIS-Shepherd~\cite{pan2025vis} & Code + image & \Checkmark & \XSolidBrush & \XSolidBrush & \Checkmark & \XSolidBrush \\
VisJudge-Bench~\cite{xie2026visjudge} & Image-only & \XSolidBrush & \XSolidBrush & \Checkmark & \Checkmark & \XSolidBrush \\
\midrule
\textbf{DashArena (ours)} & \makecell[c]{Code + image +\\interaction evidence} & \Checkmark & \Checkmark & \Checkmark & \Checkmark & \Checkmark \\
\bottomrule
\end{tabular}
\caption{Scope comparison with representative visualization generation and evaluation benchmarks. ``Controlled'' denotes ambiguity injected around known chart specifications; it differs from an open-ended analytical request with multiple valid dashboard designs.}
\label{tab:benchmark-comparison}
\end{table*}

Our final benchmark contains 234 tasks spanning 14 topic clusters and a broad range of structural complexity.
We evaluate seven mainstream LLMs and include the original human-authored dashboards as an anonymized human baseline.
To validate the judge, we invited six independent annotators to label 100 stratified candidate pairs, each with three annotations.
On the 99 evaluable pairs, DashJudge-8B obtains 79.8\% human agreement and $\kappa=0.600$.
Ablating trajectories and their execution reports lowers agreement to 71.7\%, while deterministic execution rules alone reach only 42.4\%, showing that interaction evidence is useful but insufficient without multimodal semantic judgment.
The final leaderboard is stable across Bradley--Terry, Thurstone--Mosteller, and average-win-rate aggregation.

Our contributions are:
\begin{itemize}
    \item We introduce DashArena, to our knowledge the first benchmark for open-ended, task-grounded generation of interactive analytic dashboards, together with datasets, task documents, and anonymized human baselines.
    \item We develop an interaction-grounded evaluation methodology that turns generator-authored, replayable interaction trajectories into execution evidence for multimodal pairwise judgment.
    We release the distilled open-weight judge, DashJudge-8B, as a reproducible and efficient implementation of this methodology.
    \item We conduct a comprehensive empirical study of seven frontier models and the human baseline, supported by a human calibration study, deterministic reliability metrics, aggregation diagnostics, robustness checks, and a 100-case failure audit.
\end{itemize}

\section{Related Work}

\paragraph{Visualization generation and evaluation.}
NL2VIS systems map language to visualization code, while later work extends generation to dashboards and LLM-based workflows~\cite{luo2018deepeye,narechania2020nl4dv,zhou2021table2charts,deng2022dashbot,wu2021multivision,maddigan2023chat2vis,tian2024chartgpt,ouyang2025nvagent}.
Yet benchmarks remain mostly chart-level: nvBench and nvBench~2.0 test text-to-visualization and ambiguity~\cite{luo2021nvbench,luo2025nvbench}, and VisEval and VIS-Shepherd use VLM judges for single charts~\cite{chen2024viseval,pan2025vis}.
VisJudge-Bench covers multi-view dashboards but not interaction or task completion~\cite{xie2026visjudge}.
Interactive multi-view generation has also been studied~\cite{zhao2024lightva,qiu2025smartmlvs,shen2025dashchat,shi2026nl2dashboard}, but these systems are evaluated largely through expert studies instead of scalable benchmarks.
Dashboard2Code evaluates static dashboard reconstruction~\cite{niu2026dashboard2code} but does not consider open-ended tasks or interaction.
DashArena is the first benchmark to evaluate interactive dashboards on open-ended tasks, with runtime validity, interaction evidence and visual quality all considered in the evaluation.

\paragraph{Interactive artifacts and multimodal judging.}
Design2Code targets static page reconstruction, whereas WebArena and VisualWebArena evaluate agents acting in existing sites~\cite{si2025design2code,zhou2024webarena,koh2024visualwebarena}.
MLLM-as-a-Judge motivates pairwise multimodal evaluation while documenting judge bias~\cite{chen2024mllmjudge}.
ArtifactsBench evaluates broad executable artifacts but lacks dashboard-specific data analysis and linked-view evaluation~\cite{zhang2025artifactsbench}; MiniAppBench and Vibe Code Bench depend on GUI agents exploring created artifacts~\cite{zhang2026miniappbench,tran2026vibe}.
Following GameCraft-Bench's replay-before-judging principle~\cite{luo2026gamecraft}, DashArena uses submitted trajectories to obtain reproducible interaction evidence for analytic dashboards.

\section{DashArena}
\label{sec:benchmark}

\subsection{Task Construction}

\paragraph{Collection and quality filtering.}
We crawl dashboards from Tableau Public\footnote{\url{https://public.tableau.com/app/search/vizzes/dashboard}.}, a platform for sharing interactive data visualizations, together with their metadata, page screenshots, workbook files\footnote{Workbooks are XML files that store the dashboard structure and interaction actions.}, and tabular data.
We search for dashboards and sort them by popularity, so the crawled dashboards are likely to have high community recognition.
We first remove entries that cannot provide all four artifact types, then use a VLM to retain analytical dashboards with at least two usable views and to exclude tutorials, templates, and non-analytical showcases.
We deduplicate sources by underlying-data similarity while retaining limited alternative human designs for the same data; full details are provided in supplementary material.
We collect only publicly accessible artifacts and preserve each source URL and creator attribution.
The released benchmark distinguishes project-derived tasks and metadata from original Tableau assets, which remain subject to creator rights, platform terms, and source-specific licenses; the supplementary material details provenance, redistribution, and removal procedures.

\paragraph{Task derivation.}
Claude Opus 4.6 receives all reference pages, the sampled-data schema, and interaction actions extracted from the workbook file.
It produces a structured seed containing a title, user context, one concise task brief, non-exhaustive analytical goals, and desired interaction capabilities.
The prompt forbids copying screenshot-specific KPI values and asks for analytical intent rather than chart, layout, or pixel-level reproduction.
The seed is then converted into a natural-language task document with a fixed template, appending database schema and common implementation constraints.
We inspect the resulting tasks and confirm that all analytical goals are achievable with the provided data.
The resulting 234 tasks constitute the released benchmark.
The tasks span 14 data-driven topic clusters and a broad range of structural complexity; complete distributions and construction details appear in supplementary material.

\subsection{Candidate Generation}

Each model receives the task document and data schema and must return one complete HTML document.
For comparable browser evaluation, we require ECharts\footnote{\url{https://echarts.apache.org/}.} and stable \texttt{data-test-id} attributes on interactive elements.
Tabs and multiple internal pages are allowed, and the constraints do not prescribe specific design choices.

Generation uses a continuous two-turn conversation.
After producing the HTML, the model receives the interaction-trajectory schema and generates a chain of actions.
The second turn retains the complete context, so the model can refer to the dashboard code that it has just generated.

\paragraph{Why model-authored trajectories?}
Inspired by GameCraft-Bench~\cite{luo2026gamecraft}, we deliberately ask the evaluated model to author the trajectory.
An open-ended dashboard has no canonical exploration policy: the same task may be implemented through various layout and interaction designs.
A judge that must discover these mechanisms is simultaneously an interface-exploration agent, so its failures to locate a control or reach a page would be incorrectly attributed to the candidate.
Its search policy would also introduce additional stochasticity and latency into the benchmark.
Instead, we view the trajectory as executable usage documentation, analogous to the README or demonstration test supplied with a code repository.
It communicates the artifact's intended entry points and representative workflow to the judge.

Candidate-authored evidence does not mean candidate-controlled evaluation.
The model only declares what should be attempted and what analytical change it expects; the executor verifies what actually happens, and the judge determines whether the demonstrated behavior is useful for the task.
The judge is instructed to consider the trajectory's analytical value and coverage, not only its execution success, so a short trace that selectively avoids promised views is penalized.

\subsection{Interaction Trajectory and Execution}

An interaction trajectory is a JSON object containing only an ordered \texttt{steps} array.
Each step specifies an action type, a \texttt{data-test-id} target, a natural-language analytical intent, and the expected visual change.
The schema supports setting, typing into, or clicking controls; clicking arbitrary DOM elements; clicking chart marks; and brushing chart regions.
Chart actions may optionally identify a datum.
Tab-switching DOM actions mark whether an additional page screenshot should be captured so that all pages visited during the trajectory are available to the judge.
Full schema details are provided in supplementary material.

Our Playwright executor validates the JSON schema, loads the candidate in Chromium, and captures the initial screenshot.
It inserts a fixed delay after each action and records action errors plus the identifiers of changed charts, controls, and text components by comparing serialized component signatures before and after execution.
Control signatures include values and checked states; text signatures include rendered content; chart signatures include container text, visibility and geometry, and canvas or SVG renderer metadata.
The executor captures every step for diagnostics but exposes to the judge only the screenshots of every visible page and a summary of which components changed after each action.
This is meant to reduce the judge's inference cost and memory usage while still providing sufficient visual evidence.
Local routing serves CSV data files and pinned copies of common browser libraries, preventing network instability from being counted as a model failure.

We separately report two deterministic validity measures.
A candidate is \emph{renderable} when its page loads without error, required data requests resolve, and the initial screenshot is non-blank.
It is \emph{replayable} when it is renderable, its trajectory is schema-valid, every declared action executes without error, and all required step and page screenshots are captured.
Thus, an invalid trajectory does not by itself imply that the dashboard cannot render, while replayability remains a property of the joint dashboard--trajectory submission.
These checks are not folded into the pairwise preference judgments.

\subsection{Interaction-Aware Pairwise Judge}

We use pairwise judgments rather than assign each candidate an absolute score.
A pointwise scale requires globally calibrated anchors: the same numerical rating must retain its meaning across datasets, task difficulty, and multiple valid analytical designs while combining partially competing qualities such as analytical utility, interaction, and presentation.
A rubric can make these considerations explicit and reduce scoring ambiguity, but it does not by itself determine task-invariant anchors or universally valid weights among them.
Pairwise evaluation conditions on the same task and data and asks which candidate better supports the requested analysis, avoiding cross-task scale drift.

The VLM judge compares two anonymous candidates for the same task.
Its input contains the task instructions, dataset schema, page-level screenshots, and trajectory execution reports.
The judge is prompted to consider analytical coverage, interaction, and visual clarity, and to treat execution reports as hard evidence of whether promised actions actually occurred.
The output is a structured preference (candidate A/B better or tie) and a concise rationale.

\paragraph{Human baseline.}
For each task, we present the source Tableau dashboard as an anonymous candidate without revealing its origin to the judge.
This gives an interpretable reference point for model performance.
Since our evaluation protocol requires an interaction trajectory which the original dashboard does not provide, we prompt Claude Opus 4.6 to generate a schema-valid trajectory from the screenshots and extracted interactions, and we express its execution report in the same format as model candidates.
One author manually verify that every generated trajectory is schema-valid and consistent with the interactions extracted from the original workbook; these trajectories are treated as protocol-valid rather than programmatically replayed against Tableau.

\subsection{Open-weight Judge Distillation}

Repeated proprietary VLM calls make a large leaderboard expensive and difficult to reproduce, so we distill the final judge into an open-weight VLM, Qwen3-VL-8B-Instruct.
The teacher labels 300 pairs stratified across comparison categories, of which 255 pairs from 114 tasks form the training split and 45 pairs from 20 disjoint tasks form the held-out Judge test split.
The 20 judge-test tasks are a subset of the 120 benchmark tasks absent from SFT, rather than an additional split; all 120 form the held-out leaderboard split.
For training only, we add an A/B-swapped copy of every pair and reverse the label, yielding 510 SFT examples.

We train for two epochs with LoRA rank 16, scale 32, dropout 0.05, learning rate $10^{-4}$, effective batch size 8, and a cosine schedule.
It takes approximately 50 minutes to complete training on a single 80 GB A100 GPU.

\subsection{System-Level Aggregation}

For models $i$ and $j$, the Bradley--Terry model defines
\begin{equation}
P(i \succ j)=\sigma(\theta_i-\theta_j),
\end{equation}
where $\theta_i$ is a latent ability score for model $i$ and $\sigma$ is the logistic function.
We report the centered transformation $1000+400\theta_i$ for readability.
Candidate order is randomized independently for each comparison.
Confidence intervals resample complete tasks, preserving dependence among all candidates evaluated on the same instruction and dataset.


\section{Experiments}
\label{sec:experiments}

\subsection{Experimental Setup}

We evaluate GPT-5.5, Claude Opus 4.6, GLM-5.2, Gemini 3.5 Flash, DeepSeek V4 Pro, Kimi K2.7 Code, and Grok 4.3, with one candidate per model and task.
The main leaderboard uses the 120 held-out evaluation tasks, none of which appears in DashJudge-8B's SFT data.
After excluding comparisons with incomplete candidate evidence and invalid or non-directional judge outputs, 3,325 comparisons remain for aggregation; full accounting appears in the supplement.

To create a human calibration set, we sample 100 pairs from 50 tasks, comprising 75 model--model and 25 human--model comparisons.
To reduce trivial one-sided non-replayable cases, the sampling strategy requires 90 pairs to be both replayable, and the remaining 10 pairs to be replayable vs. non-replayable.
The 100 pairs are not used for DashJudge-8B training or leaderboard aggregation.
We recruit six independent annotators who have experience with dashboard design.
The annotators each label 50 pairs, giving three independent annotations per pair, without access to model identities or judge outputs.
Annotators inspect the task, field schema, page screenshots, and interaction trajectory evidence, and choose whether A or B is better, or tied.
The annotation set contains 52 unanimous pairs, 46 directional two-to-one majorities, one two-to-one tie majority, and one three-way A--B--tie split with no majority; Fleiss' $\kappa$ among annotators is 0.384.
Only 8 of 300 annotations (2.7\%) select tie.
We exclude the no-majority pair and denote the remaining 99 pairs as $\mathcal{C}$ and use them to evaluate judge agreement and positional consistency.

\subsection{Is DashJudge-8B Reliable?}

On 45 held-out teacher-test pairs from 20 disjoint tasks, DashJudge-8B reproduces the proprietary teacher label with 88.9\% accuracy.
Table~\ref{tab:judge-human} evaluates human alignment on $\mathcal{C}$.
DashJudge-8B improves over its base model by 9.1 agreement points and raises Cohen's $\kappa$ from 0.419 to 0.600, while its human agreement is comparable to the proprietary teacher.

\begin{center}
\centering
\small
\setlength{\tabcolsep}{4pt}
\begin{tabular}{lcc}
\toprule
Judge & Agreement (\%) & $\kappa$ \\
\midrule
Base (Qwen3-VL-8B-Instruct) & 70.7 & 0.419 \\
Proprietary teacher (Claude Opus 4.6) & 78.8 & 0.590 \\
\textbf{DashJudge-8B} & \textbf{79.8} & \textbf{0.600} \\
\bottomrule
\end{tabular}
\captionof{table}{Agreement with human majority preferences on $\mathcal{C}$.}
\label{tab:judge-human}
\end{center}

Human ambiguity explains a substantial part of the remaining disagreement.
Using all available DashJudge-8B predictions, agreement is 92.3\% on the 52 unanimous pairs but only 66.0\% on the 47 pairs with a two-to-one majority.

We also replace the judge with Gemini 3.1 Pro Preview on pairs in $\mathcal{C}$ and a balanced 560-pair sample from the held-out leaderboard tasks.
Gemini is less aligned with human majorities than the Claude teacher (68.7\% versus 78.8\%), indicating that proprietary judges are not interchangeable at the pair level.
Nevertheless, Gemini and DashJudge-8B agree on 81.6\% of labels, and their induced rankings have Kendall's $\tau=0.857$ and preserve 26 of 28 model-pair orders, providing evidence that the model-level result is not specific to the Claude teacher; full results appear in the supplement.

\subsection{Does Interaction-Aware Evaluation Matter?}

We rerun DashJudge-8B on $\mathcal{C}$ after removing both trajectories and execution reports while leaving the task and screenshots unchanged.
As Table~\ref{tab:interaction-ablation} shows, full interaction evidence has an 8.1-point higher agreement and a 0.159 higher $\kappa$.

\begin{center}
\small
\begin{tabular}{lcc}
\toprule
Judge input & Agreement (\%) & $\kappa$ \\
\midrule
Task + screenshots & 71.7 & 0.441 \\
+ interaction evidence & \textbf{79.8} & \textbf{0.600} \\
\bottomrule
\end{tabular}
\captionof{table}{Interaction-aware Judge ablation on $\mathcal{C}$.}
\label{tab:interaction-ablation}
\end{center}

We also compare deterministic rules that rank candidates lexicographically by renderability, replayability, step success, and changed-component evidence.
Renderability and replayability checks alone predict a tie for 89 of 99 pairs and obtain only 8.1\% human agreement.
The full rule reaches 42.4\% agreement and $\kappa=0.095$, far below DashJudge-8B's 79.8\% and $\kappa=0.600$ on the same 99 pairs.
Thus both interaction evidence and multimodal semantic judgment are necessary for reliable pairwise evaluation.

\subsection{Do Model-Authored Trajectories Test Their Dashboards?}

We audit 100 renderable model candidates sampled from 100 distinct tasks and balanced across the seven models.
We replay the trajectory of each candidate.
Per-page control coverage is the macro-average fraction of logical controls used on each interactive page, and the stricter page-level criterion requires every interactive page to reach at least 50\% coverage and every static page to be visited.
We separately measure whether executed control actions change downstream chart data or text content, excluding the acted-on element itself.

\begin{center}
\small
\setlength{\tabcolsep}{3.5pt}
\begin{tabular}{lrr}
\toprule
Audit measure & Mean & 95\% CI \\
\midrule
Valid trajectory targets & 99.8 & 99.3--100.0 \\
Per-page control coverage & 80.8 & 75.0--86.3 \\
All pages visited & 96.0 & 89.0--98.0 \\
Authored-control downstream response & 90.1 & 84.3--95.2 \\
Unauthored-control downstream response & 86.0 & 76.5--95.8 \\
\bottomrule
\end{tabular}
\captionof{table}{Trajectory-coverage audit over 100 candidates (percent; candidate-bootstrap intervals).}
\label{tab:trajectory-coverage}
\end{center}

The audit shows that model-authored trajectories usually target valid controls and reach most pages, but they do not exhaustively exercise every interactive element.
The gap between the 80.8\% per-page coverage and a theoretical 100\% is largely due to the single-trajectory length limit and the common practice of generating multiple parallel controls (e.g., several side-by-side filters) but using only one in the trajectory.
We consider this a reasonable design choice.

To prevent style-only changes from receiving content-change credit, we hash ECharts runtime data and compare text content while excluding the acted-on target itself.
We then independently probe each unique native parameter control from a clean page state, using up to two valid alternatives and an explicit apply button when present.
Of 362 controls, 246 appear in the authored trajectory and 116 do not; 233 and 107, respectively, admit a safely inferred alternative value.
All 340 probes execute without runtime error.
Strict downstream responses occur for 90.1\% of authored and 86.0\% of unauthored controls.
Restricting the comparison to selects yields 92.6\% versus 91.8\%.
We then manually inspect all four candidates that fail to visit every page.
DashJudge-8B explicitly identifies missing page evidence or the associated analytical omission for all four cases.
Therefore, we find no evidence that omitted controls fail more often than controls selected for the walkthrough, nor any case with clear evidence that a model selectively concealed a known failure.

We further inspect all 841 authored steps against replay status, before/after screenshots, exact control values, rendered text, ECharts runtime data, and generated HTML when attribution is ambiguous.
The executor yields attributable step-level evidence for 818 of 841 actions (97.3\%).
Among these actions, 763 (93.3\%) agree with the declared intent and 55 (6.7\%) expose a candidate or trajectory error.
The remaining 23 actions (2.7\% of all steps) are conservatively abstained: all execute without error, but ECharts canvas replay cannot verify that the requested datum received the click.
Across the existing pairwise rationales, DashJudge-8B identifies the same error, or its failed prerequisite, for 53 of the 55 confirmed error steps and for 29 of the 31 affected candidates.
This audit shows that authored trajectories usually describe substantive behavior and tend to expose, rather than conceal, their own failures.
Full step-level decisions and counterfactual chart-handler probes are provided in the supplement.

\subsection{Main Leaderboard}

Figure~\ref{fig:main-results} reports the final pairwise results.
GPT-5.5 ranks first, although its CI overlaps the human baseline's.
The human baseline and GLM-5.2 occupy the next two positions with overlapping uncertainty, while Claude, DeepSeek, Gemini, and Kimi form a broad middle tier.
Grok 4.3 is clearly last.

\begin{figure*}[t]
    \centering
    \includegraphics[width=\textwidth]{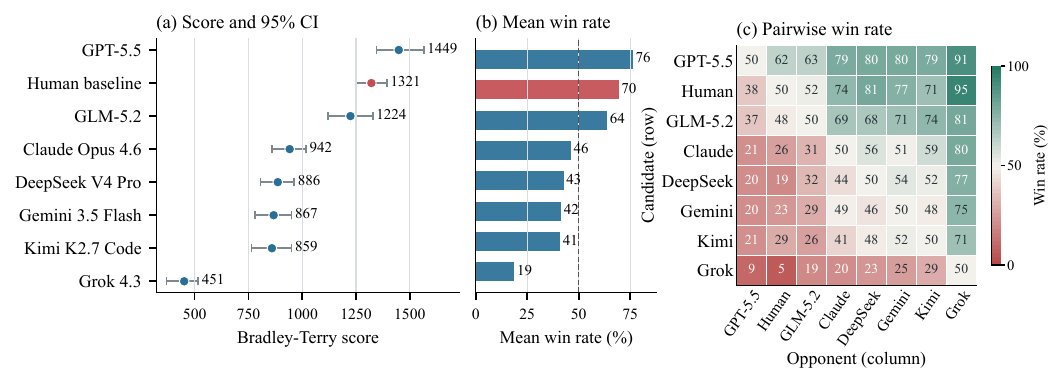}
    \caption{Main leaderboard on 120 held-out tasks from 3,325 directional DashJudge-8B preferences. (a) Scaled Bradley--Terry scores with task-bootstrap 95\% confidence intervals; numbers denote point estimates. (b) Macro-average win rates against the other systems. (c) Pairwise win rates (\%), where each cell gives the row system's win rate over the column system.}
    \label{fig:main-results}
\end{figure*}

Table~\ref{tab:reliability} complements preference-based quality with browser-derived reliability.
No model exceeds an 86\% render rate or a 74\% replay rate, showing that high pairwise quality does not eliminate basic runtime and interaction failures.
GPT-5.5 leads both measures, while GLM's high preference rank coexists with relatively low functional reliability.

\begin{center}
\centering
\small
\setlength{\tabcolsep}{4pt}
\begin{tabular}{lrr}
\toprule
System & Render rate (\%) & Replay rate (\%) \\
\midrule
GPT-5.5 & \textbf{85.8} & \textbf{73.3} \\
GLM-5.2 & 74.2 & 54.2 \\
Claude Opus 4.6 & 82.5 & 72.5 \\
DeepSeek V4 Pro & 75.0 & 54.2 \\
Gemini 3.5 Flash & 83.3 & 62.5 \\
Kimi K2.7 Code & 63.3 & 47.5 \\
Grok 4.3 & 84.2 & 67.5 \\
\bottomrule
\end{tabular}
\captionof{table}{Deterministic measures on the leaderboard tasks.}
\label{tab:reliability}
\end{center}

\subsection{Qualitative Failure Analysis}

We audit 100 candidates from 86 tasks, balanced across seven models: 70 retrieved by render/replay failure signals and 30 execution-clean controls.

\begin{center}
\centering
\small
\setlength{\tabcolsep}{4pt}
\begin{tabular}{lrr}
\toprule
Primary mechanism & Signal-enriched & Exec.-clean \\
\midrule
Construction/runtime & 18 (26.9) & 0 (0.0) \\
Interaction fulfillment & 17 (25.4) & 3 (14.3) \\
Data computation/binding & 15 (22.4) & 7 (33.3) \\
Presentation/readability & 11 (16.4) & 6 (28.6) \\
Analytical fulfillment & 6 (9.0) & 5 (23.8) \\
\bottomrule
\end{tabular}
\captionof{table}{Primary causes among confirmed model-attributed defects in the signal-enriched ($n=67$) and execution-clean candidates ($n=21$) strata; cells show count (\%).}
\label{tab:failure-analysis}
\end{center}

67 of 70 (95.7\%) signal-enriched candidates contain a confirmed model-attributed defect.
Of the other three, one is a benign, stable refresh/sort operation and two are isolated chart-click delivery or capture ambiguities despite implemented ECharts handlers.
We exclude these three only from the defect-mechanism counts in Table~\ref{tab:failure-analysis}.
Separately, 21 of 30 execution-clean candidates contain semantic, analytical, interaction, or visual defects that deterministic browser checks are not designed to detect.
These candidates contain no construction/runtime failures; instead, manual inspection attributes 18 of the 21 defects to data computation or binding, presentation, and analytical coverage, showing why successful loading and trajectory replay cannot establish dashboard quality.
For example, these candidates concatenate overlapping components, treat a filtered-away comparison group as zero and report a false value, or replace a requested county choropleth with a non-geographic rectangular heatmap.

Runtime failures include persistent loaders, unavailable map dependencies, and code exceptions; manual inspection additionally finds malformed schema binding, unsupported field derivation, and invalid aggregation.
Other dashboards render credible shells but omit requested analysis, expose unwired controls, author infeasible interaction paths, or compress labels and series beyond readability.

\subsection{DashJudge-8B Training Analysis}

\paragraph{Data scaling.}
We train nested subsets with 0, 64, 128, and 255 independent teacher pairs, retaining swapped augmentation and all other hyperparameters.
As shown in Table~\ref{tab:data-scaling}, 64 pairs recover most of the gain over the base model, while 255 pairs achieve the best agreement and positional consistency, indicating high data efficiency but not yet saturation.

\begin{center}
\small
\setlength{\tabcolsep}{4pt}
\begin{tabular}{rrrr}
\toprule
Pairs & Agreement & $\kappa$ & Mirror cons. \\
\midrule
0 & 70.7 & 0.419 & -- \\
64 & 76.8 & 0.539 & 91.9 \\
128 & 76.8 & 0.540 & 89.9 \\
255 & \textbf{79.8} & \textbf{0.600} & \textbf{93.9} \\
\bottomrule
\end{tabular}
\captionof{table}{DashJudge-8B data scaling on $\mathcal{C}$ (\%).}
\label{tab:data-scaling}
\end{center}

\paragraph{A/B swap augmentation.}
We isolate A/B augmentation by training the 255 original pairs for four epochs, matching the 1,020 sample exposures of two epochs over 510 augmented examples.
Without swapping, human agreement decreases by only 1.0 point, but mirror consistency drops from 93.9\% to 83.8\% and first-position win rate shifts from 51.0\% to 42.9\%.
The 10.1-point consistency loss has a task-bootstrap 95\% interval of $[1.0,19.2]$ points, confirming that swap augmentation primarily removes positional sensitivity.

\subsection{Leaderboard Robustness}


We first test sensitivity to the aggregation rule by comparing Bradley--Terry with Thurstone--Mosteller~\cite{thurstone2017law, mosteller1951remarks} and average-win-rate aggregation.
Replacing the logistic link with the Gaussian link of Thurstone--Mosteller, or simply ranking models by their observed average win rates, produces exactly the same eight-model order as BT.
Thus, the reported order is robust to the choice of link function and to the use of a latent variable model.

We next change how tasks contribute to the fit.
Giving every task equal weight preserves the full ordering, while equal topic weighting swaps only the overlapping DeepSeek and Gemini positions.
After removing one topic at a time, the resulting rankings retain $\tau\geq0.857$ with the full ranking, and no model moves by more than two positions.
These checks show that neither tasks with more valid comparisons nor any single topic cluster determines the main capability tiers.

Finally, we repeatedly sample tasks without replacement, refit BT, and compare each induced order with the full 120-task evaluation ranking using Kendall's $\tau$.
Mean $\tau$ rises from 0.810 with 25 tasks and 0.867 with 50 tasks to 0.945 with 100 tasks; among 300 samples of 100 tasks, both the top model and top-three set are recovered in 100\%.
The analysis indicates that the leading tier is stable, while small studies remain insufficient for resolving the closely spaced middle models.

\section{Discussion}

\paragraph{Human baseline.}
Task seeds abstract each reference dashboard's high-level intent without requiring the reference to satisfy the resulting instruction.
Models optimize that instruction directly, whereas the reference predates it; the baseline therefore remains a practical anchor rather than an oracle.

\paragraph{Human preference diversity.}
Open-ended dashboards permit legitimate differences in how viewers trade analytical coverage, interaction, and visual clarity.
We use one overall pairwise label rather than prescribe universal weights for these qualities, so the leaderboard represents aggregate preference under our protocol.
Of 100 human-labeled pairs, 99 yield a majority and 52 yield unanimous preference, showing that the evaluation protocol is sufficiently clear to produce a strong consensus.
Estimating population-level preference distributions would require many more judgments per pair, broader annotator sampling, and stronger expertise and attention controls, substantially increasing cost and quality-control difficulty.
We therefore leave distributional modeling outside the benchmark scope.
A preliminary analysis is included in the supplementary material.

\paragraph{Limitations and future work.}
First, Tableau Public carries platform-specific biases, leaving private and other visualization ecosystems outside the current scope.
Second, model-authored trajectories enable deterministic replay without relying on an exploratory agent, but restrict direct evidence to the submitted walkthrough.
The protocol verifies observable execution yet latent numerical or data errors may remain undetected.
Third, budget constraints limit our experiments to model-only, single-conversation generation rather than iterative coding-agent harnesses such as Codex or Claude Code.
This bounds the leaderboard, not the protocol, which can evaluate any served web dashboard accompanied by the same structured trajectory.
Future work can broaden task sources, commission a matched-protocol human baseline, and include agent harnesses in the leaderboard.

\section{Conclusion}

We introduced DashArena, a 234-task benchmark for open-ended generation of interactive analytic dashboards.
Its protocol combines model-authored interaction trajectories, browser execution, and multimodal pairwise judgment to evaluate candidates.
Human calibration, ablations, audits, and aggregation diagnostics provide evidence for DashJudge-8B as a reproducible evaluator; across seven frontier models and a human baseline, the benchmark reveals failures behind visually plausible output.

\bibliography{aaai2027}


\end{document}